\documentclass[sigconf,nonacm]{acmart}

\AtBeginDocument{%
  }

\usepackage{algorithm}
\usepackage{algpseudocode}
\usepackage{graphicx}
\usepackage{subcaption}
\usepackage{enumitem}
\usepackage{booktabs}
\usepackage{multirow}
\usepackage{tabularx}
\usepackage{amsmath}
\usepackage{bm}

\renewcommand{\shortauthors}{Agrawal and Hayes}
\begin{document}

\title{GIST: Multimodal Knowledge Extraction and Spatial Grounding via Intelligent Semantic Topology}

\author{Shivendra Agrawal}
\email{shivendra.agrawal@colorado.edu}
\affiliation{%
  \institution{University of Colorado Boulder}
  \city{Boulder}
  \state{Colorado}
  \country{USA}
}

\author{Bradley Hayes}
\email{bradley.hayes@colorado.edu}
\affiliation{%
  \institution{University of Colorado Boulder}
  \city{Boulder}
  \state{Colorado}
  \country{USA}
}

\begin{abstract}
Navigating and searching complex, densely packed environments such as retail stores, warehouses, libraries, and hospitals poses a great spatial grounding challenge for both humans and embodied AI. In these environments, dense visual features become stale quickly given the quasi-static nature of items, and long-tail semantic distributions pose a challenge for dedicated computer vision methods. Vision-Language Models (VLMs) have emerged as a popular paradigm to help assistive systems and robots understand and navigate semantically-rich spaces, yet they still struggle with spatial grounding in dense, cluttered, quasi-static environments. We present \emph{GIST} (Grounded Intelligent Semantic Topology), a multimodal knowledge extraction pipeline that transforms a consumer-grade mobile point cloud into a semantically annotated navigation topology. Our architecture distills the scene into a 2D occupancy map, extracts its topological layout, and overlays a lightweight semantic layer via intelligent keyframe and semantic selection. We demonstrate the versatility of this structured, semantically-rich spatial knowledge through critical downstream Human-AI interaction tasks: (1) an intent-driven \emph{Semantic Search} engine that actively infers categorical alternatives and zones when exact matches fail; (2) a one-shot \emph{Semantic Localizer} achieving 1.04\,m top-5 mean translation error; (3) a \emph{Zone Classification} module that segments the walkable floor plan into high-level semantic regions; and (4) a \emph{Visually-Grounded Instruction Generator} that synthesizes optimal paths into egocentric, landmark-rich natural language routing. Evaluated via an independent multi-criteria LLM evaluation protocol, GIST outperforms navigation instruction generation baselines. Finally, an in-situ formative evaluation ($N=5$) yields an 80\% navigation success rate relying solely on verbal cues, validating the system's capacity for universal design.
\end{abstract}

\ccsdesc[500]{Computing methodologies~Spatial and physical reasoning}
\ccsdesc[500]{Computing methodologies~Natural language generation}
\ccsdesc[500]{Human-centered computing~Accessibility}
\ccsdesc[500]{Human-centered computing~Ubiquitous and mobile computing}
\ccsdesc[500]{Human-centered computing~Interactive systems and tools}

\keywords{Navigation, Topology, Semantic Map, Multimodal Data, Universal Design, Interactive systems}

\begin{teaserfigure}
\centering
\includegraphics[width=\textwidth]{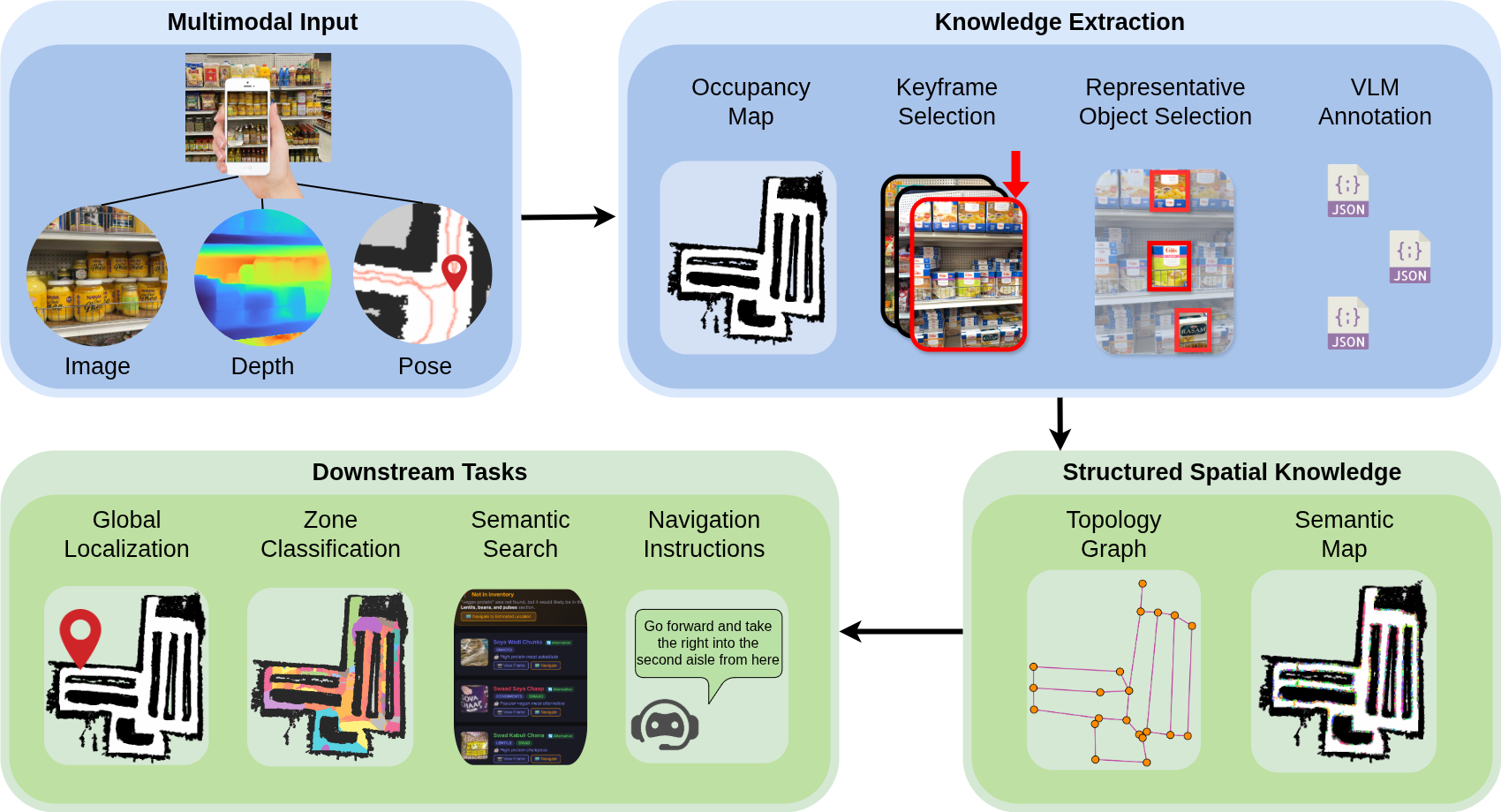}
\Description{A block diagram showing mobile RGB-D data processed through object extraction and a Vision-Language Model to create a semantic topology map, which subsequently powers intent-aware search, semantic localization, and natural language routing.}
\caption{The \emph{GIST} Multimodal Knowledge Extraction Architecture. Raw multimodal inputs (RGB-D and mobile odometry) are distilled via intelligent keyframe selection, representative object selection, and VLM labeling into Structured Spatial Knowledge. This shared representation enables robust downstream Human-AI interaction and autonomous system tasks, including intent-aware semantic search, global pose localization, and spatially grounded natural language routing.}
\label{fig:system_arch}
\end{teaserfigure}

\maketitle

\section{Introduction}

Navigating and searching dense, cluttered environments poses a significant spatial grounding challenge for both humans and autonomous embodied AI. Conventional mapping techniques capture geometric free space, but they do not represent the semantic structure people rely on for search, orientation, and route understanding \cite{nav2_amcl}. 

In dense quasi-static environments, navigation depends heavily on semantic landmarks and topological transitions rather than solely metric commands. Prior work in spatial cognition shows that people interpret routes more effectively when directions are chunked into meaningful segments and anchored to recognizable landmarks \cite{lovelace1999elements,klatzky1998allocentric}. For example, `walk past the lentils on your left, then turn right' is more actionable and lower-burden than `walk 8.6 meters forward and then turn right.' 

These spaces are inherently \emph{quasi-static}—environments where the spatial distribution, orientation, and quantity of objects evolve over extended timescales, such as fluctuating product inventory on grocery shelves or moving books in a library. Furthermore, as demonstrated by our 3,500 sq. ft. international grocery store testbed, they exhibit extreme long-tail distributions of culturally specific, visually similar items that induce severe perceptual aliasing. Recent Vision-and-Language Navigation (VLN) frameworks perform well in benchmarked and synthetic domains~\cite{krantz2020vlnce, chen2024mapgpt}, but consistently underperform in real-world settings~\cite{aghzal2025evaluating, wang2025rethinking, windecker2025navitrace} where target semantics may be partially occluded, visually repetitive, or absent from the immediate view.

Recent instruction generation frameworks, such as NavComposer~\cite{he2025navcomposer}, are an important step toward scalable data generation, but they are optimized for visual sequence modeling rather than human-centered spatial communication. In dense environments, this leads to directions that often depend on transient scene content and under-specify stable route geometry, definitive turn structure, and durable landmarks. 

To bridge this gap, we introduce \emph{GIST} (\textbf{G}rounded \textbf{I}ntelligent \textbf{S}emantic \textbf{T}opology), an end-to-end framework for extracting a semantically grounded navigation representation from a single consumer mobile scan (Figure~\ref{fig:system_arch}). GIST converts RGB-D data and odometry into a 2D occupancy map, derives a topology graph over walkable space, and anchors semantically labeled products to that graph through keyframe selection, representative object selection, and VLM-based annotation. The resulting semantic topology separates deterministic geometric structure from higher-level semantic reasoning while keeping both in a shared coordinate frame. 

This structured representation supports critical Human-AI interaction tasks: (1) \textit{Intent-Driven Search \& Missing Goal Estimation} to estimate zone locations when exact matches are not available on the semantic map; (2) \textit{Semantic Localization}, a one-shot localizer estimating an agent's discrete pose $(x, y, \theta)$ from a single smartphone image using pure text-embeddings; (3) \textit{Zone Classification} over walkable space; and (4) \textit{Visually-Grounded Spatial Routing} translating spatial paths into view-independent, natural language directions.  
Our key contribution is the shared intermediate representation that supports search, localization, and communication via a coordinated, human-centered approach.

Across 15 real-world scenarios, we demonstrate that GIST's explicit topological structures significantly outperform raw RGB sequence baselines in multi-criteria LLM evaluations, particularly as route complexity increases. Finally, a formative in-situ ecological probe ($N=5$) yields an 80\% navigation success rate using generated verbal cues alone, providing encouraging initial evidence of real-world utility.

\section{Related Work}
\emph{GIST} facilitates several critical human-AI and embodied AI tasks; the following subsections present related work foundational to these domains. 

\subsection{Vision-and-Language Navigation (VLN)}
VLN requires an embodied agent to interpret natural language instructions and navigate 3D environments solely from visual observations~\cite{anderson2018vln}. Early benchmarks operated in discrete graph-based settings where agents teleport between panoramic viewpoints~\cite{qi2020reverie}. While VLN-CE~\cite{krantz2020vlnce} pushed toward continuous environments requiring low-level motor control, these systems still assume high-quality instructions are already provided.

Automated instruction generation has emerged to scale VLN training data. Early speaker models used LSTM-based networks to translate panoramic sequences into directions~\cite{fried2018speaker}, lacking spatial specificity. Recent models like NavRAG~\cite{wang2025navrag} utilize scene description trees to provide fine-grained context. Goal-conditioned generation models like GoViG~\cite{wu2025govig} attempt to produce instructions from only a first-person observation and a goal image. NavComposer~\cite{he2025navcomposer} generates instructions from RGB image sequences combined with discrete action traces. However, because sequence-to-sequence methods bypass explicit topological reasoning, they struggle to infer path geometry from images alone. This results in instructions that reference visually salient but transient objects (\textit{e.g.}, red crates, shopping carts) and cannot reliably specify definitive turn sequences, exact metric distances, or macro-level navigation instructions.

\subsection{Localization in Quasi-Static Environments}
Particle filter-based localization methods (e.g., AMCL~\cite{foxKLDSamplingAdaptiveParticle2001, nav2_amcl}) remain standard for onboard robotics. However, their reliance on static geometric maps makes them notoriously brittle in quasi-static environments like warehouses and retail stores, where local semantics fluctuate even when global geometry remains constant~\cite{yinSurveyGlobalLiDAR2024}.

Semantic SLAM systems~\cite{goswamiEfficientRealTimeLocalization2023a, zimmermanLongTermLocalizationUsing2023} often assume landmark stability, an assumption that fails in real-world quasi-static environments where objects frequently shift or fluctuate in quantity. Recent probabilistic approaches like \emph{ShelfAware}~\cite{agrawal2025shelfaware} model shelf semantics as statistical distributions over object counts to handle these changes. However, they rely strictly on low-level visual features, which often fail to distinguish semantically distinct but visually similar items (e.g., bags of sugar, salt, or flour). Deep visual localization methods~\cite{dong2025reloc3r, loiseau2025alligat0r} and implicit neural representations~\cite{kuangIRMCLImplicitRepresentationBased2023} demonstrate impressive accuracy but require server-grade compute, precluding real-time use on wearable edge devices. GIST overcomes these limitations by utilizing one-shot, text-embedding-based semantics to enable instant global localization on lightweight hardware.

\subsection{Assistive Guidance Systems}
Navigating large retail spaces or warehouses presents a critical challenge for people with visual impairments, involving macro-navigation in the locomotor space and micro-navigation in the haptic space~\cite{gharpure2008robot}.

\textbf{Macro-Navigation and Product Search:} Recent work focuses largely on navigation inside the store~\cite{kulyukin2010accessible,kulyukin2005robocart,kulyukin2006ergonomics}. Solutions that rely on environmental augmentation, such as RFID tags~\cite{lopez-de-ipinaBlindShoppingEnablingAccessible2011} or Bluetooth beacons, introduce high maintenance overhead and barriers to adoption. For product identification, existing techniques using fixed-class object detectors~\cite{feng2020research} or barcode scanning~\cite{lookout, nicholson2009shoptalk} are impractical for the sheer volume of long-tail retail items~\cite{NielsenIQ}. While researchers have explored conversational agents~\cite{kamikubo2024we, kaniwa2024chitchatguide} and the ``last few meters'' way-finding problem~\cite{saha2019closing}, these systems often still rely on BLE beacons, underscoring a critical need for uninstrumented, scalable semantic search.

\textbf{Manipulation Guidance and Social Dynamics:} Once a user is in the vicinity of a product, manipulation guidance helps them find regions of interest~\cite{vazquez2014assisted, bonani2018my}. Crowdsourced assistance like Be My Eyes~\cite{bemyeyes} or VizWiz~\cite{bigham2010vizwiz} relies on human availability. Some researchers have focused on the product retrieval sub-task by providing verbal manipulation guidance for product retrieval off the shelf \cite{agrawalShelfHelpEmpoweringHumans2023, agrawal2023assistive}. However, these systems inherently assume the user is already perfectly localized in front of the correct shelf~\cite{zientara2017third}. GIST bridges this gap by creating intermediate spatial knowledge that focuses on solving the prerequisite locomotor navigation problem by searching for desired goals and estimating the current pose of the system.

\section{Multimodal Knowledge Extraction}

Our offline pipeline converts consumer mobile RGB-D data into a reusable semantic topology without manual annotation.

\subsection{Mobile Data to 2D Occupancy}
We captured a real-world, 3,500 sq ft international grocery store using a consumer mobile device equipped with LiDAR. A 14.5-minute traversal yielded 52,006 RGB-D frames, which were subsampled $6\times$ to 8,668 frames ($\sim$10 fps) alongside 6-DoF ARKit visual-inertial odometry. To ensure lightweight processing and downstream scalability, we explicitly avoid computationally heavy classical SLAM backends. Instead, we project a 2D occupancy grid (0.05\,m/pixel) by applying a height-slice to the raw mobile point cloud.

\subsection{Informative Keyframe \& Object Extraction}
To reduce multimodal input processing and annotation cost, we first perform keyframe selection.
Each full RGB frame is embedded via DINOv3 (\texttt{vitb16}, 768-dim). A sequential cosine filter retains only frames whose similarity to the previously accepted keyframe falls below a 0.85 threshold, condensing the sequence to 660 visually distinct keyframes (a 92.4\% frame reduction) to eliminate redundancy.

Each keyframe is processed by a YOLOv9 model fine-tuned on the SKU-110K dataset \cite{goldman2019dense} for dense shelf products. Because tightly-packed retail shelves generate an extreme density of bounding boxes for repetitive products, we filter the detections to maintain only crops that maximize image quality and semantic diversity. Specifically, we compute the Laplacian variance (sharpness) for every crop and employ a \emph{Feature-Space Strategy}: from each frame, we select the absolute sharpest crop alongside the crop exhibiting the maximum DINOv3 cosine distance to ensure maximum visual diversity. Finally, fresh produce (fruits and vegetables), which are rarely placed on traditional shelves, are extracted via a secondary pre-trained YOLOv9-COCO pass.

\subsection{Semantic Map}
\label{sec:semantic}
To optimize VLM inference throughput and rate-limiting, these representative crops are arranged into $5 \times 5$ mosaic grid images (39 grids total). These mosaics are processed by a unified VLM (\texttt{gemini-3-flash-preview}) to extract the product's \texttt{name}, \texttt{brand}, \texttt{packaging\_type}, and \texttt{category}. The VLM is also tasked to classify the crop as blurry or sharp. We only use the sharp crops for downstream tasks (such as localization, zone classification, and as landmarks for natural language generation) that are sensitive to classification errors. 

\textbf{Object Position Refinement:} Object map positions are computed via depth-based unprojection, mapping the bounding box center through the synchronized depth map into the ARKit world coordinate frame. Specifically, given the homogeneous pixel coordinates $\tilde{\bm{u}} = [u, v, 1]^\top$ at the bounding box's \emph{median} depth $Z_c$, the map position $\bm{X}_w$ is calculated as:
\begin{equation}
\bm{X}_w = \bm{t}_{wc} + Z_c \bm{R}_{wc} \bm{K}^{-1} \tilde{\bm{u}}
\end{equation}
where $\bm{K}$ represents the camera intrinsic matrix, and $\bm{R}_{wc}$ and $\bm{t}_{wc}$ denote the camera-to-world rotation and translation, respectively.
Because consumer depth sensors frequently suffer from noise near shelf edges due to quasi-static products, these raw unprojected points occasionally fall inside structural obstacles. To correct this, we implement Bresenham’s line algorithm \cite{koopman1987bresenham} along the camera line-of-sight to drag-or-pull the object by casting a ray from the camera origin through the product toward the nearest occupied pixel and free space boundary (the shelf wall). This refinement snaps products to plausible shelf-adjacent locations, improving usefulness for search and route planning. 

\subsection{Topology}
\label{sec:topology}
We construct a navigation graph via morphological skeletonization of the occupancy map~\cite{zhang1984fast}. Skeleton pixels form a pixel-level graph from which we extract junctions (pixel degree $\geq 3$). Turn points are detected using a sliding-window angular change detector (turn threshold=60\textdegree). Junctions and turns within a small threshold radius are merged via hierarchical clustering, and short dead-end spurs are pruned. Bresenham line-of-sight checks on the high-resolution occupancy grid ensure all edges are traversable. Non-traversable edges are removed. To bind semantic products to the geometric topology graph, we compute the perpendicular projection of the product coordinate onto the nearest graph edge. We dynamically insert a \emph{virtual node} at this precise projection point during pathfinding. This reduces reach distance dramatically, ensuring the extracted A* path travels deep into the correct aisle and arrives within 0.1--0.5\,m of the physical target. This also allows our system to figure out the side of aisle the product is on. 

\begin{figure}[b!]
\centering
\includegraphics[width=0.85\linewidth, trim=0 1.5cm 0 0,clip]{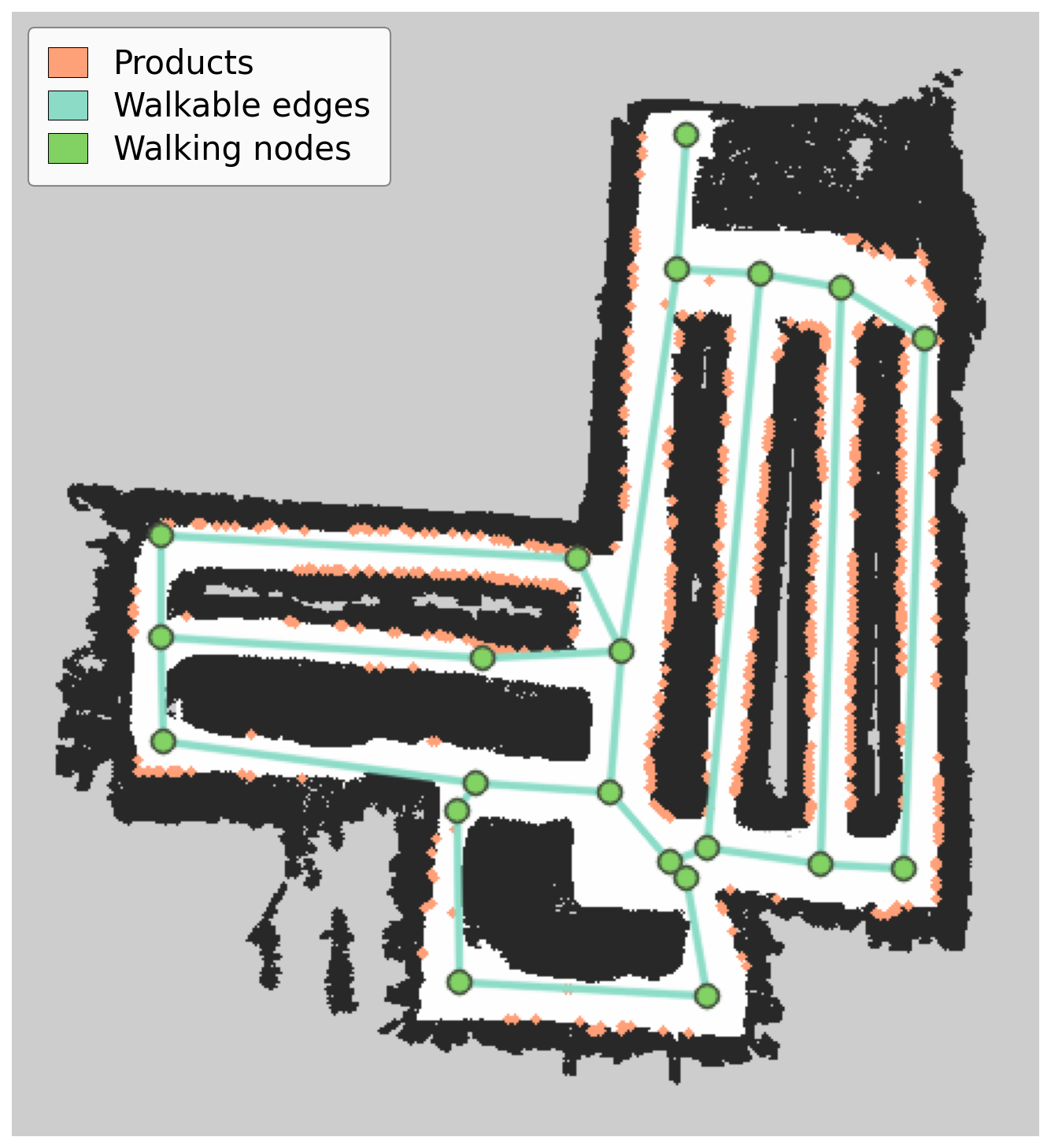}
\Description{A 2D top-down floor plan showing a green skeletonized navigation graph with teal edges. Hundreds of orange dots representing localized products are clustered along the edges of the path.}
\caption{GIST Semantic Topology: The skeletonization-derived graph providing walking nodes (green) and traversable edges (teal), overlaid with localized products (orange).}
\label{fig:topology-map}
\vspace{-2mm}
\end{figure}

\section{Downstream Human-AI Tasks}

The extracted semantic topology serves as a shared foundation for multiple downstream interaction tasks. To ensure architectural simplicity and consistency, all generative reasoning tasks are powered by the same \texttt{gemini-3-flash-preview} foundation model.

\subsection{Zone Classification}

\begin{figure}[t!]
\centering
\includegraphics[width=0.85\linewidth, trim=0 1.5cm 0 0,clip]{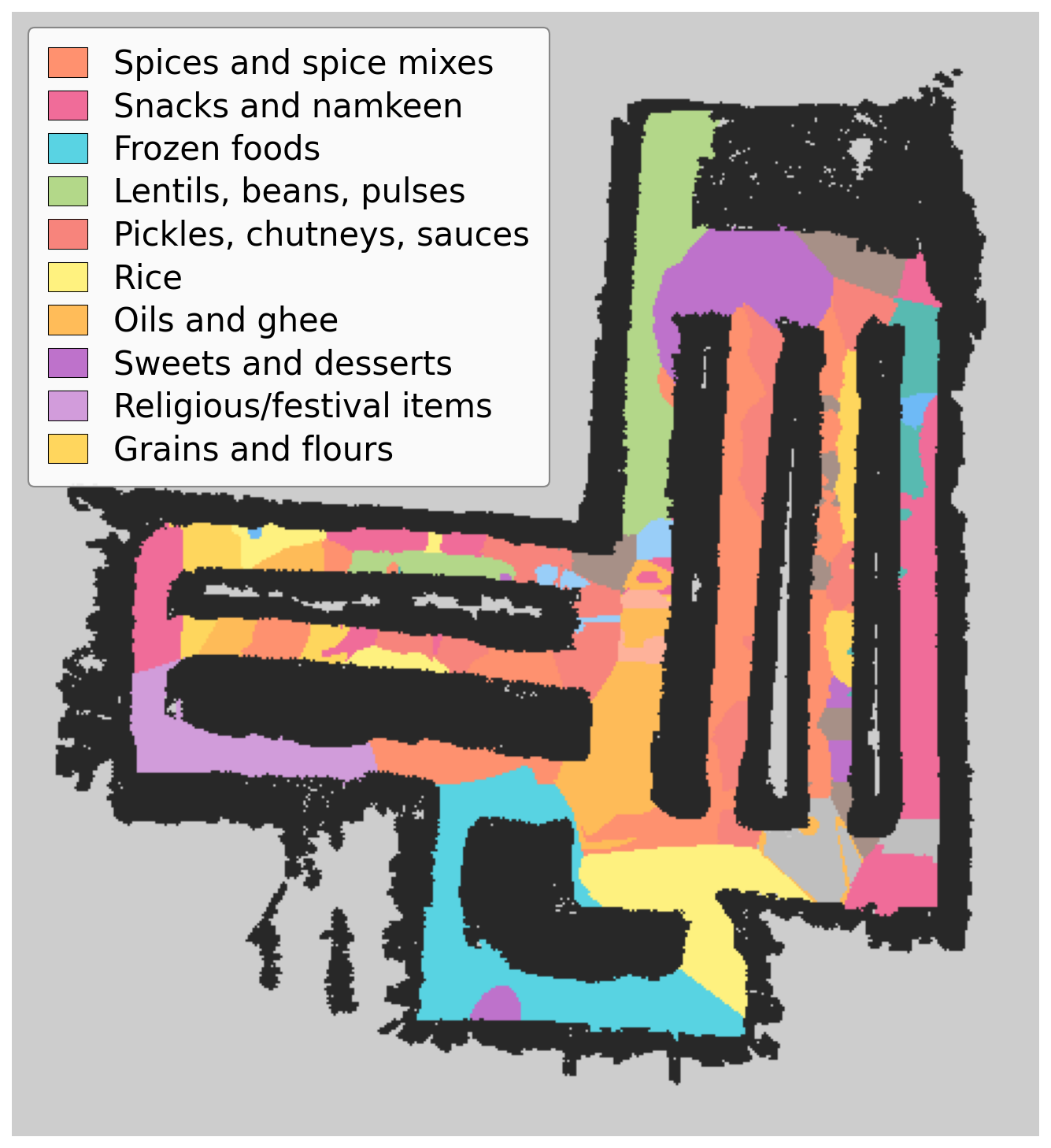}
\Description{A 2D top-down floor plan where the walkable space is colored in various shaded regions. Each color corresponds to a specific grocery zone, such as Lentils or Spices, as indicated by a legend.}
\caption{Semantic Zone Classification: Free-space pixels are assigned to 18 specific semantic zones via KD-tree voting over the localized product positions (legend displays the top 10 zones by area).}
\label{fig:zone-map}
\vspace{-2mm}
\end{figure}

To provide high-level semantic context, all discovered products are classified into 18 specific zones (e.g., \texttt{Lentils}, \texttt{Spices}) via a single VLM prompt analyzing the unique string metadata. A KD-Tree is constructed from all mapped product coordinates. For every free-space pixel in the occupancy map, an inverse-distance-squared weighted vote of the $k=5$ nearest products assigns a continuous semantic zone overlay to the walkable floor plan (Figure~\ref{fig:zone-map}). 

This continuous overlay exposes higher level semantic organization not explicitly encoded in the floor plan.
For instance, the system generated a dense clustering of heterogeneous categories near the top-left aisle region. Ground-truth verification of the raw frames confirmed this region corresponds to the store's dedicated ``Organic'' section. This demonstrates the architecture's capacity to autonomously discover emergent macro-structures from micro-level semantic distributions.

\subsection{Intent-Aware Search \& Category Estimation}
\begin{figure}[t!]
\centering
\begin{subfigure}{0.48\columnwidth}
\centering
\includegraphics[width=\linewidth, height=8cm, keepaspectratio=false]{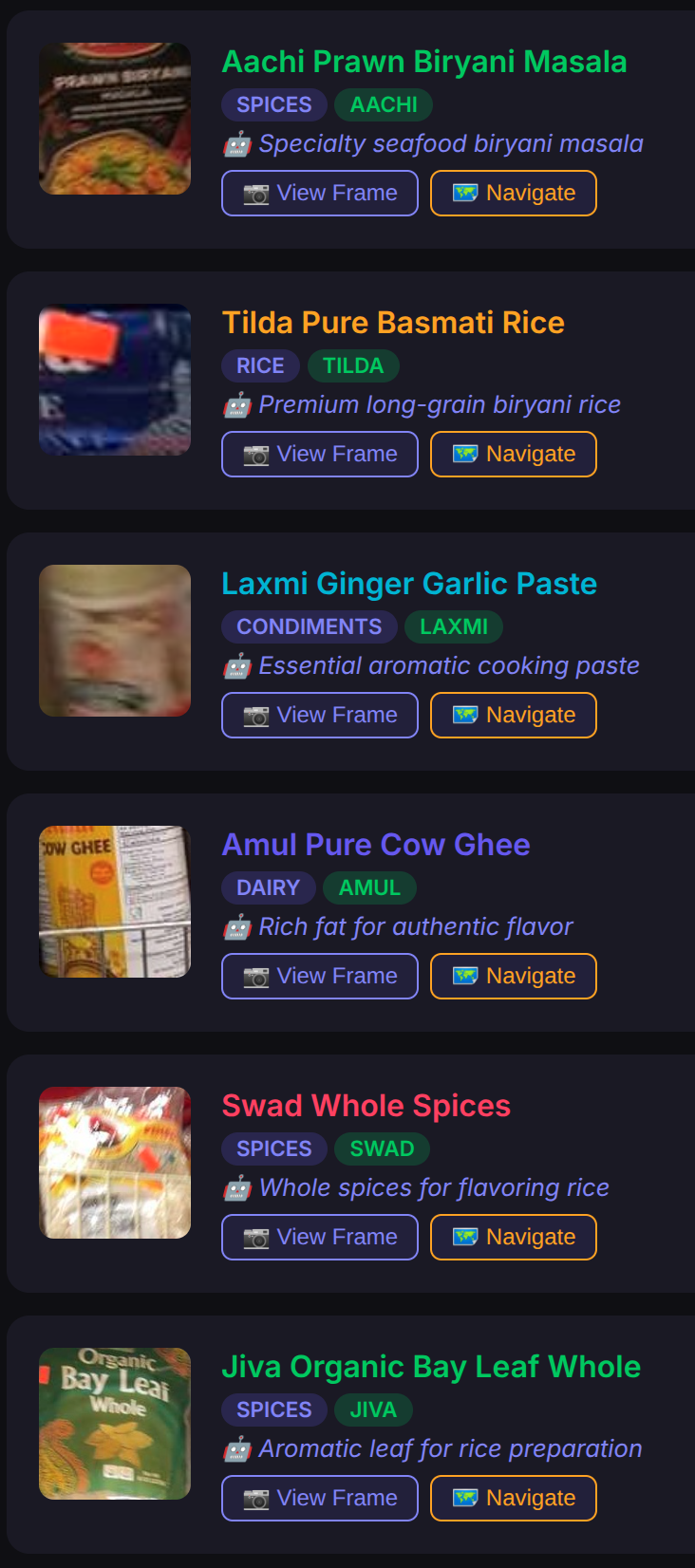}
\end{subfigure}
\hfill
\begin{subfigure}{0.48\columnwidth}
\centering
\includegraphics[width=\linewidth, height=8cm, keepaspectratio=false]{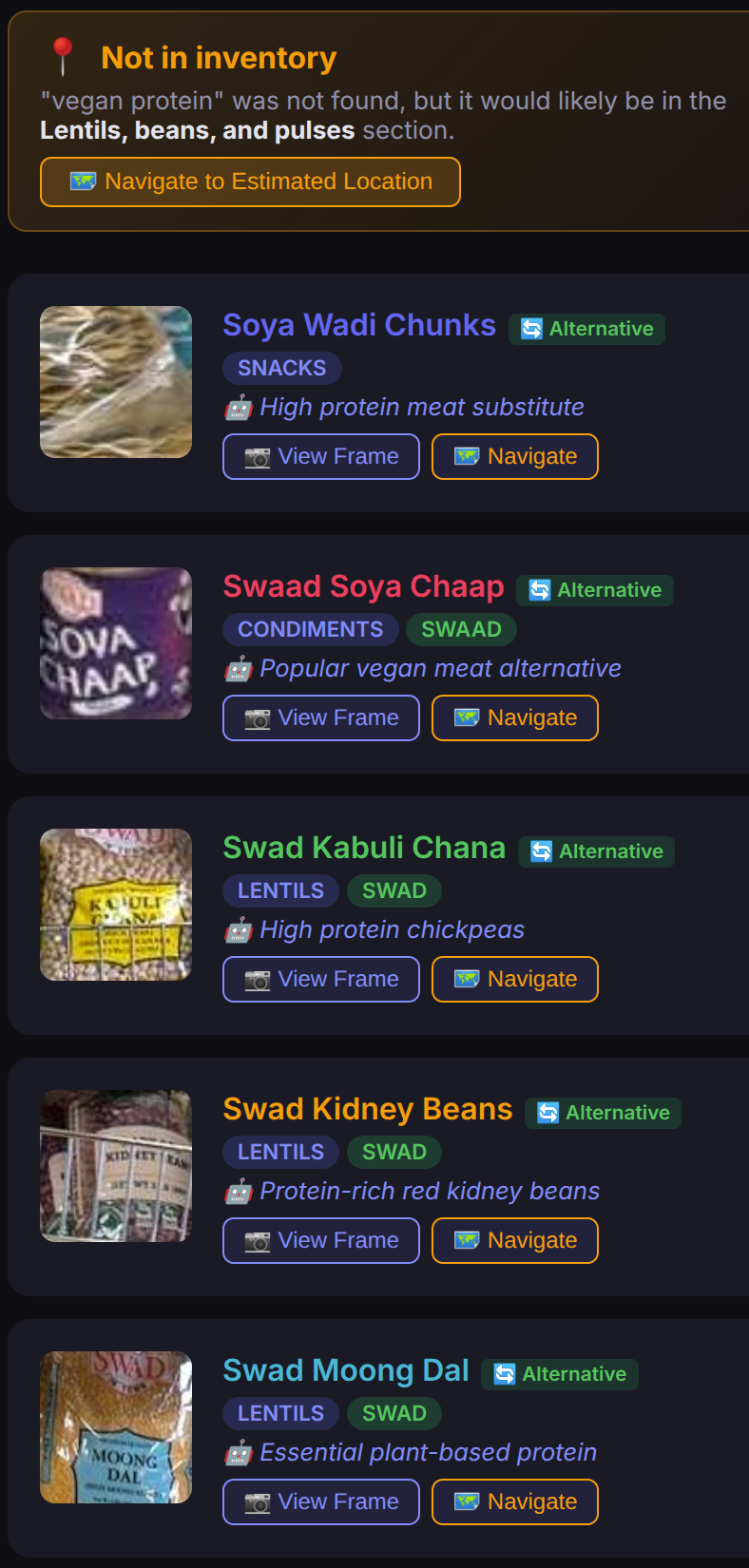}
\end{subfigure}
\Description{Two screenshots of a mobile web interface. The left screen shows search results for a multi-item recipe, listing distinct ingredients. The right screen shows an abstract query for vegan protein resolving to a general Lentils and Beans zone.}
\caption{Intent-aware search via the Gemini-powered web interface. \textbf{Left:} A multi-item recipe query (``biryani ingredients'') is decomposed into distinct physical targets. \textbf{Right:} An abstract query (``vegan protein'') demonstrating how the VLM infers alternative semantic categories to enable robust spatial estimation.}
\label{fig:search}
\end{figure}

Standard keyword matching is often inadequate for long-tail, culturally unique, and dense inventory where spelling variations and linguistic substitutions are common. To resolve this, we leverage the VLM to perform intent-aware semantic matching against our structured product map, deployed via an interactive web interface. As shown in Figure~\ref{fig:search} (Right), abstract queries like ``vegan protein'' return not just exact matches, but items categorized as \texttt{Alternative} or \texttt{Related}, accompanied by generative reasoning explaining the selection. The system explicitly predicts the encompassing semantic zone (e.g., \texttt{Lentils, beans, and pulses}). By calculating the spatial median of the largest contiguous pixel cluster for that zone on the semantic map, it routes the user to the correct general aisle rather than failing outright if a specific product is unmapped.

This graceful fallback via \emph{Category Estimation} is critical in practice because physical perception via mobile scanning inherently suffers from field-of-view (FOV) clipping (e.g., bottom-shelf items obscured by narrow aisles) and dynamic occlusions during the initial mapping phase. Furthermore, this intent-aware architecture natively supports complex, multi-goal queries. As seen in the ``biryani ingredients'' query (Figure~\ref{fig:search}, Left), the VLM effectively decomposes a prepared dish into its key distinct components---biryani spices, basmati rice, garlic paste, and ghee---and simultaneously locates or estimates the topological zones for all required items, laying the groundwork for optimized multi-stop path planning.

\subsection{One-Shot Text-Based Semantic Localizer}
To support navigation without external infrastructure or computationally heavy visual feature matching, we introduce a semantic localizer that estimates user pose $(x, y, \theta)$ from a single smartphone image (Algorithm~\ref{alg:semantic_loc}). This capability is also crucial for autonomous systems to localize in quasi-static environments.

Offline, we discretize the occupancy grid into 0.5m cells with 8 orientation bins (45\textdegree{} each). We perform camera FOV raycasting (utilizing 20 rays per pose) against the semantic map. The collected product metadata along each ray is converted to a text string, sorted alphabetically to ensure permutation invariance, concatenated, and encoded via DistilBERT \cite{sanh2019distilbert} into a 768-d $L_2$-normalized vector, generating the semantic map $\mathcal{M}_{text}$. We cache the expected semantics for each of the discrete poses for faster lookup. Online, a single query image is processed via YOLOv9 to extract crops, labeled via the VLM, and embedded via DistilBERT. Cosine similarity against $\mathcal{M}_{text}$ returns the top-$k$ ranked pose hypotheses.

While our current implementation operates purely on semantics and yields encouraging zero-shot results, this semantic-based feature can be naturally fused with continuous onboard sensor data (e.g., accelerometers and depth sensors) to provide the high-frequency tracking required for autonomous robotic systems to safely operate in these spaces.

\begin{algorithm}[h!]
\caption{One-Shot Text-Based Semantic Localization}
\label{alg:semantic_loc}
\small
\textbf{Input:} Query RGB image $I_q$, Precomputed Pose Map $\mathcal{M}_{text}$ \\
\textbf{Output:} Estimated Global Pose Hypothesis $H = (x, y, \theta)$
\begin{algorithmic}[1]
\State \textbf{Phase 1: Online Query Processing}
\State $C_{crops} \gets \text{YOLOv9}(I_q)$ \Comment{Extract product bounding boxes}
\State $L_{raw} \gets \text{VLM}(\text{crops}) \to \{\text{brand}, \text{category}\}$
\State $L_{deduplicated} \gets \text{SortAlphabetically}(\text{Unique}(L_{raw}))$
\State $E_q \gets \text{DistilBERT}(L_{deduplicated})$ \Comment{text embedding}
\State \textbf{Phase 2: Semantic Matching}
\State $\mathcal{S} \gets \emptyset$
\For{each free-space pose $p_i$ and precomputed embedding $E_i \in \mathcal{M}_{text}$}
    \State $\text{score} \gets \text{CosineSimilarity}(E_q, E_i)$
    \State $\mathcal{S} \gets \mathcal{S} \cup \{(p_i, \text{score})\}$
\EndFor
\State \Return $\text{argmax}_{k}(\mathcal{S})$ \Comment{Return top-$k$ ranked poses}
\end{algorithmic}
\end{algorithm}

\subsection{Visually-Grounded Spatial Routing}
We compute the optimal A* path across the topology graph. Path segments are logically chunked by analyzing node headings; angle changes $>30^\circ$ generate discrete turn commands, while collinear segments are merged into forward distances. 

Rather than prompting an LLM with coordinates alone, GIST provides explicitly grounded visual context. The 2D occupancy map is rendered as a color image containing the topology graph, the A* path, and start/goal markers (Figure~\ref{fig:topology-map}). To provide semantic landmarks without hallucinatory visual clutter, we perform Bresenham line-of-sight rasterization. From the midpoints of the A* path, all products within a 50-pixel (2.5m) radius are tested for visibility against the occupancy grid's static obstacles. Visible products are deduplicated by category, and their relative side (left/right) is calculated via the cross-product against the path heading vector. This annotated map image, alongside the sequential route segments and validated landmarks, is passed to the VLM to generate the final egocentric routing instructions, explicitly prohibiting the use of cardinal directions.

\section{Evaluation Setup}

The full suite of pipeline parameters enabling exact reproducibility is detailed in Table~\ref{tab:parameters}. To comprehensively evaluate the system, we defined 15 navigation scenarios across the 3,500 sq ft store (Table~\ref{tab:scenarios}), ranging from simple to complex, multi-turn paths.

\begin{table}[h]
\centering
\caption{Implementation \& Reproducibility Parameters}
\label{tab:parameters}
\small
\begin{tabularx}{\columnwidth}{@{}lX@{}}
\toprule
\textbf{Component} & \textbf{Hyperparameter / Configuration} \\
\midrule
\textbf{Data Capture} & Consumer LiDAR Smartphone (iPhone) \\
\textbf{RGB-D Input} & 8,668 frames, 960$\times$720 RGB, 256$\times$192 Depth \\
\textbf{2D Map} & 0.05m/px resolution \\
\textbf{Keyframe Filter} & DINOv3 (\texttt{vitb16-pretrain}), Cosine $\leq 0.85$ \\
\textbf{Product Detection} & YOLOv9 (SKU-110K), conf $\geq 0.25$ \\
\textbf{Produce Detect.} & YOLOv9 (COCO) \\
\textbf{VLM Engine} & \texttt{gemini-3-flash-preview} (Temp: 0.3) \\
\textbf{VLM Batching} & $5 \times 5$ representative mosaic grids \\
\textbf{Skeletonization} & Zhang-Suen morphological thinning \\
\textbf{Localization} & \texttt{distilbert-base-uncased} (768-d), 0.5m grid, 45$^\circ$ bins, 60$^\circ$ FOV, 6m range, 20 rays \\
\bottomrule
\end{tabularx}
\end{table}

\begin{table}[h]
\centering
\caption{The 15 Benchmark Navigation Scenarios used for evaluation, spanning culturally specific product categories.}
\label{tab:scenarios}
\resizebox{\columnwidth}{!}{%
\begin{tabular}{@{}llcc | llcc@{}}
\toprule
\textbf{ID} & \textbf{Target Product} & \textbf{Dist.} & \textbf{Turns} & \textbf{ID} & \textbf{Target Product} & \textbf{Dist.} & \textbf{Turns} \\
\midrule
s01 & Laxmi Chana Dal         & 2.9m  & 0 & s14 & Badshah Chatpat Masala  & 8.6m  & 3 \\
s09 & Swad Blackeye Peas      & 4.5m  & 0 & s15 & McVitie's Digestives    & 8.6m  & 3 \\
s10 & Mother's Recipe Pickle  & 4.7m  & 1 & s04 & Badshah Pav Bhaji       & 11.2m & 0 \\
s11 & Patak's Chili Pickle    & 5.0m  & 1 & s05 & Swad Soya Wadi          & 13.5m & 1 \\
s02 & Aachi Biryani Masala    & 6.5m  & 1 & s06 & Swad Organic Urad Dal   & 14.2m & 1 \\
s12 & Pani Puri Kit           & 6.5m  & 2 & s07 & Guntur Chilli Powder    & 17.1m & 2 \\
s03 & Haldiram's Moong Dal    & 8.6m  & 1 & s08 & Deep Aloo Paratha       & 19.4m & 2 \\
s13 & Parle Krackjack         & 8.6m  & 3 &     &                         &       &   \\
\bottomrule
\end{tabular}%
}
\end{table}

\subsection{Baseline Configurations and Prompts}
To evaluate the contribution of semantic topology to route generation, we compare GIST against baselines that reflect visual-sequence generation, naive coordinate prompting, and ablations of GIST itself:
\begin{itemize}[leftmargin=*]
    \item \textbf{NavComposer~\cite{he2025navcomposer}}: Provided with a sequence of first-person RGB trajectory frames sampled along the A* path, paired with discrete action traces (e.g., \emph{``Step 1: walk forward, Step 2: turn left''}), strictly replicating the model's native input formulation. It received no topological maps, zone labels, or metric distances.
    \item \textbf{Naive Gemini}: Provided with a text-only prompt containing the target product name, total metric distance, store dimensions, and the allocentric cardinal direction from the start position.
    \item \textbf{GIST (Text)}: Provided with the full array of parsed topological route segments, turn counts, total walking distance, and zone classifications, but completely deprived of the visual map image.
    \item \textbf{GIST (Visual)}: Provided with the fully annotated topological map image (graph, A* path, start/goal markers, scale bar) and the route segments, total distance, and landmarks (category and relative side).
\end{itemize}

\section{Results and Discussion}

\subsection{Semantic Localization Accuracy}
To validate the One-Shot Semantic Localizer (Algorithm~\ref{alg:semantic_loc}), we evaluated 20 random shelf-facing frames from the mapping traversal. The true ARKit 6-DoF pose serves as ground truth.

\begin{table}[h]
\centering
\caption{Semantic localization translation error.}
\label{tab:localization}
\small
\begin{tabular}{lcc}
\toprule
\textbf{Metric ($k$)} & \textbf{All 20 Frames} & \textbf{Correct Zone (80\%)} \\
\midrule
Top-1 Error & 3.78 $\pm$ 3.76m & 2.71 $\pm$ 2.38m \\
Top-3 Error & 2.05 $\pm$ 1.87m & 1.35 $\pm$ 0.91m \\
Top-5 Error & \textbf{1.45 $\pm$ 1.29m} & \textbf{1.04 $\pm$ 0.80m} \\
\bottomrule
\end{tabular}
\vspace{-3mm}
\end{table}

As shown in Table~\ref{tab:localization}, the localizer achieves 80\% zone-level accuracy on 20 randomly sampled shelf-facing frames. Among correctly-zoned frames, expanding the search to top-5 hypotheses reduces the mean error to just 1.04\,m. This indicates that the correct global pose is often recovered within a small candidate set without visual feature matching, using only the distribution of observed textual semantics. 

\begin{figure}[h]
\centering
\includegraphics[width=0.7\columnwidth]{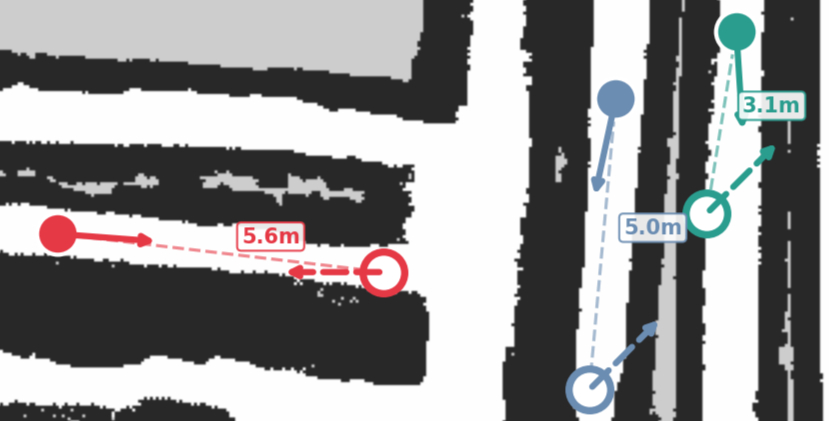}
\Description{A zoomed-in section of the 2D map showing semantic aliasing. Solid icons represent the true camera poses, while hollow icons represent the predicted poses, which are incorrectly mirrored on the opposite side of the aisle looking at the same shelf.}
\caption{Semantic aliasing in localization. Ground Truth poses (solid) and Top-1 Predicted poses (hollow) frequently exhibit large translation errors while observing the exact same semantic targets from ``mirrored'' viewpoints.}
\label{fig:mirrored}
\end{figure}

Qualitative analysis of the highest-error cases reveals a clear failure mode due to \emph{semantic aliasing} (Figure~\ref{fig:mirrored}). Mirrored or repeated viewpoints can expose nearly identical visible product semantics. 
This inflates the raw Top-1 metric translation error even when the retrieved hypothesis is semantically plausible. Accordingly, we view the one-shot localizer as a strong global initialization module within a probabilistic sensor-fusion framework (e.g., a Semantic Particle Filter), where temporal odometry and depth data can rapidly resolve these symmetric ambiguities (rather than as a complete standalone tracker).

\subsection{Independent Multi-Criteria LLM Evaluation}
Prior work \cite{zhao2021evaluation} has shown that reference-based metrics such as BLEU and ROUGE are poorly aligned with the characteristics that matter for grounded route following, while agent-centric metrics do not directly evaluate instruction quality for human users. We therefore assess route instructions using five theory-grounded criteria that target communication quality rather than lexical overlap, using the \textit{llm-as-a-judge} paradigm \cite{gu2024survey, zheng2023judging} across all 15 scenarios.
\begin{itemize}[leftmargin=*, nosep]
    \item \textbf{Egocentric Clarity:} Ensuring body-relative directions (left/right) rather than allocentric global frames \cite{klatzky1998allocentric}.
    \item \textbf{Landmark Utility:} Strategic use of structural, permanent semantics rather than transient clutter \cite{sorrows1999nature}.
    \item \textbf{Cognitive Load:} Synthesizing instructions into concise, logically chunked memory blocks \cite{sweller1991evidence}.
    \item \textbf{Safety \& Completeness:} Unambiguously specifying turns and defining definitive termination states \cite{loomis2001navigating}.
    \item \textbf{Universal Design:} Creating view-independent directions that do not rely on visual pointing UIs \cite{giudice2018navigating}. 
\end{itemize}

\begin{figure}[h]
\centering
\includegraphics[width=\columnwidth]{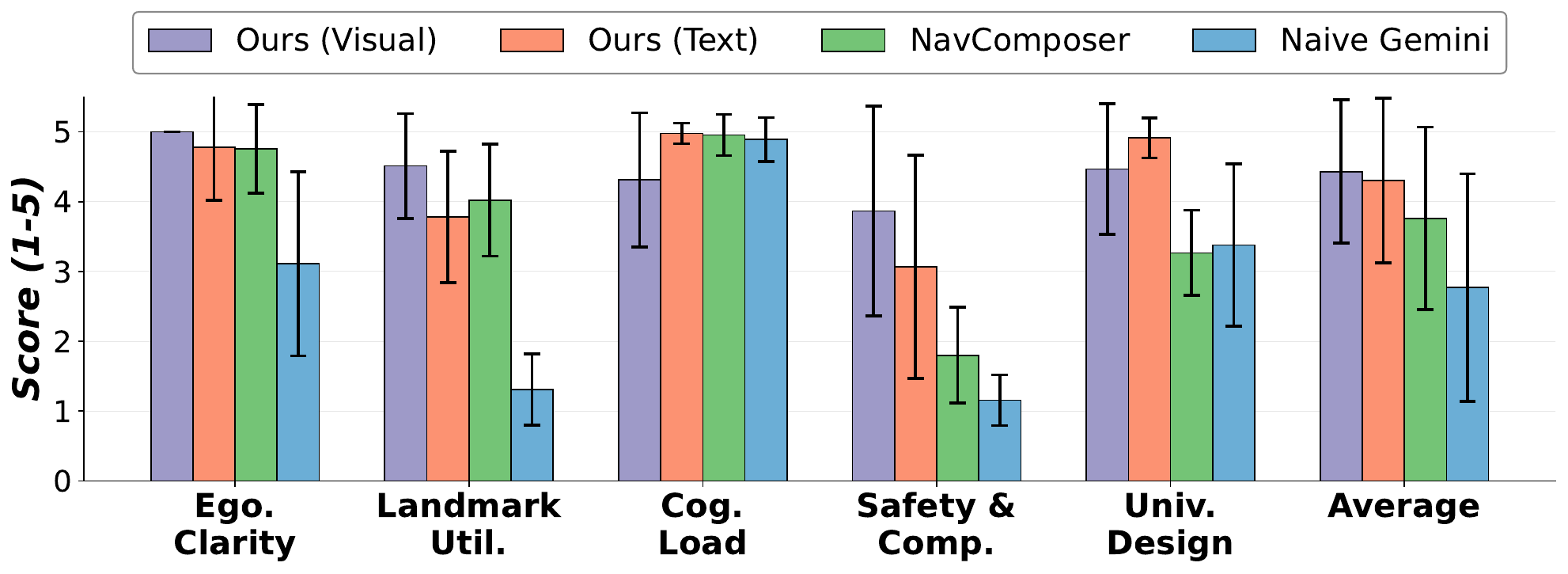}
\Description{A bar chart comparing GIST, NavComposer, and Naive Gemini across five criteria: Egocentric Clarity, Landmark Utility, Cognitive Load, Safety, and Universal Design. GIST scores the highest overall, notably achieving a perfect score in Egocentric Clarity.}
\caption{Multi-Criteria Evaluation. GIST achieves high Egocentric Clarity and overall superiority.}
\label{fig:eval_plot}
\end{figure}

As shown in Figure~\ref{fig:eval_plot}, \textbf{GIST (Visual)} achieves the highest overall score (4.43/5) including a top score on Egocentric Clarity (5/5). Providing the VLM with an annotated map image substantially eliminates cardinal direction leakage. NavComposer scores lowest on Safety \& Completeness (1.80/5) consistent with the difficulty of inferring stable route geometry from RGB frames alone. 
As demonstrated in Table~\ref{tab:examples}, NavComposer over-indexes on visual artifacts (``shopping cart'', ``yellow wall'') which may move or be non-distinct. In contrast, GIST grounds its instructions in mapped semantic structure (e.g., ``passing Canned Goods'').

\begin{table*}[t]
\centering
\caption{Representative instructions generated for two scenarios. NavComposer relies on sequential RGB frames, referencing transient objects, whereas GIST leverages the Semantic Topology for permanent, structural landmarks.}
\label{tab:examples}
\small
\begin{tabular}{p{0.12\textwidth} p{0.44\textwidth} p{0.37\textwidth}}
\toprule
\textbf{Method} & \textbf{Mother's Recipe Pickle} & \textbf{Swad Blackeye Peas} \\
\midrule
\textbf{GIST (Visual)} & Walk straight ahead for about 3 meters and then turn left into the first aisle. Continue for 2 meters, passing the Canned Goods on your right side. The Mother's Recipe Pickle is located just ahead. & Walk straight for about 11 meters, passing the Condiments on your right after 7 meters. Turn right into the first aisle and walk 2 meters further, passing Personal Care on your left to find the Swad Blackeye Peas. \\
\addlinespace
\textbf{NavComposer~\cite{he2025navcomposer}} & Walk forward past the shelves of bagged lentils and beans until you reach the shopping cart. Turn left toward the yellow wall, and you will find the Mother's Recipe Pickle on the shelving unit to your right. & Walk straight ahead past the shelving unit on your left featuring religious statues and bags of Swad beans. Continue past the red crates to the end-cap display of Heinz beans and coconut milk to find the Swad Blackeye Peas. \\
\addlinespace
\textbf{Naive Gemini} & Walk straight ahead from the entrance for about five meters. You will find the Mother's Recipe Pickle directly in front of you. & Walk straight ahead into the store for about 4.5 meters. You will find the Swad Blackeye Peas directly in front of you. \\
\bottomrule
\end{tabular}
\end{table*}

\subsection{Difficulty \& Visual Ablation Study}

As routes become longer (Table~\ref{tab:difficulty}), the Naive coordinate baseline suffers severe degradation (dropping to 2.73). In contrast, GIST maintains its instructional quality (4.59 on long routes). Because the semantic topology explicitly encodes graph structure, the model can logically chunk (and communicate) multi-turn pathways more reliably as path complexity increases.

\begin{table}[h]
  \centering
  \caption{Evaluation Scores by Route Difficulty. GIST improves its performance on complex routes.}
  \label{tab:difficulty}
  \small
  \begin{tabular}{@{}lccc@{}}
  \toprule
  \textbf{Difficulty} & \textbf{GIST} & \textbf{NavComp.} & \textbf{Naive} \\
  \midrule
  Short ($<5m$) & \textbf{4.31} & 3.64 & 3.27 \\
  Medium ($<10m$) & \textbf{4.10} & 3.82 & 2.58 \\
  Long ($>10m$)  & \textbf{4.59} & 3.75 & 2.73 \\
  \bottomrule
  \end{tabular}
\end{table}

To isolate the contribution of the visual inputs, we conducted an ablation study (Table~\ref{tab:visual-ablation}). Surprisingly, providing partial visual context—either by removing the background occupancy map (\textbf{Topology Only}, 4.55) or removing the explicit graph lines (\textbf{Map Only}, 4.49), maintained and even slightly improved the overall score compared to the full composite image (\textbf{GIST Visual}, 4.43). This suggests that reducing visual clutter aids VLM attention, provided the model receives adequate spatial grounding through either abstract path geometry or physical occupancy boundaries. However, visual structure alone is insufficient without semantic context. To demonstrate this, we evaluated \textbf{Naive Gemini + Map}, which provides the VLM with a bare occupancy map (start and goal markers only) but no topological text or product names. While the map preserves egocentric clarity, its Landmark Utility (2.89) and Safety (3.00) collapse. Ultimately, all GIST (visual) variants outperform the \textbf{Text-Only Ablation} (4.30), proving that pairing explicit semantic text with lightweight visual spatial grounding yields the safest, most human-centric routing.

\begin{table}[h]
  \centering
  \caption{Visual Ablation Study: Topology vs. Map inputs.}
  \label{tab:visual-ablation}
  \small
  \begin{tabular}{@{}lcccccc@{}}
  \toprule
  \textbf{Condition} & \textbf{Ego} & \textbf{Land} & \textbf{CogL} & \textbf{Safe} & \textbf{Univ} & \textbf{AVG} \\
  \midrule
  Topo Only & \textbf{5.00} & \textbf{4.60} & 4.56 & \textbf{4.00} & \textbf{4.60} & \textbf{4.55} \\
  Map Only & 4.87 & 4.62 & 4.33 & 3.98 & 4.64 & 4.49 \\
  GIST (Visual) & \textbf{5.00} & 4.51 & 4.31 & 3.87 & 4.47 & 4.43 \\
  \midrule
  Text-Only Abl. & 4.78 & 3.78 & \textbf{4.98} & 3.07 & 4.91 & 4.30 \\
  Naive + Map & 4.93 & 2.89 & 4.56 & 3.00 & 4.24 & 3.92 \\
  \bottomrule
  \end{tabular}
\end{table}

\subsection{Formative Ecological Probe}
To obtain an initial real-world signal, we conducted a formative in-situ ecological probe with $N=5$ graduate student evaluators in the target grocery store. Evaluators were assigned two navigation tasks sequentially, starting from the entrance. Importantly, to isolate and stress-test the efficacy of our generated spatial grounding, the application's visual path and product overlays were intentionally disabled. Evaluators relied solely on the semantic search results and generated verbal instructions. 

\begin{table}[h]
\centering
\caption{Formative Probe Results ($N=5$). Evaluators navigated relying solely on generated verbal instructions.}
\label{tab:user_study}
\small
\begin{tabular}{llcl}
\toprule
\textbf{Evaluator} & \textbf{Target Product} & \textbf{Time (s)} & \textbf{Outcome} \\
\midrule
\multirow{2}{*}{E1} & Dahi             & 75s & Success \\
                    & Marigold Biscuit & 35s & Success \\
\midrule
\multirow{2}{*}{E2} & Parle G          & 47s & Success \\
                    & Schezwan Sauce   & 32s & Success \\
\midrule
\multirow{2}{*}{E3} & Okra             & 26s & Success \\
                    & Ghee             & 93s & Success (1 wrong turn) \\
\midrule
\multirow{2}{*}{E4} & Ragi             & N/A & Failed (FOV Clip) \\
                    & Gud (Jaggery)    & N/A & Failed (FOV Clip) \\
\midrule
\multirow{2}{*}{E5} & Rajma            & 15s & Success \\
                    & Frozen Chapatti  & 64s & Success (1 wrong turn) \\
\bottomrule
\end{tabular}
\end{table}

\textbf{Quantitative Results:} Evaluators achieved an \textbf{80\% success rate} (8/10 tasks) with an average time-to-find of 48.4 seconds for successful trials (Table~\ref{tab:user_study}), defined via self-reporting upon reaching the physical destination. Notably, the system proved resilient to real-world inventory drift; in two instances, evaluators successfully navigated to the correct semantic coordinates even when the exact physical item was out of stock or substituted with a brand-equivalent. The two failed trials stemmed from a specific mapping limitation: both target items (\textit{Ragi} and \textit{Gud}) were located on the absolute bottom shelf, falling entirely below the field of view of the chest-level mobile scan. Furthermore, because these specific items were isolated from the surrounding shelves, the \emph{Zone-Level Estimation} fallback was unable to route the user to a proxy location. Finally, the observed minor route deviations (e.g., E3, E5) occurred primarily when evaluators preemptively ignored final instructions, relying instead on their own spatial priors.

\textbf{Qualitative Feedback:} Likert scale responses confirmed high subjective usability. Evaluators explicitly marked ``Agree'' or ``Strongly Agree'' that the search feature was useful, the instructions were easy to understand, and the guidance felt natural. Feedback also highlighted the nuanced challenges of egocentric routing: evaluators noted that their starting orientation directly dictates the validity of relative instructions. 

Despite this, the 80\% success rate using verbal cues alone provides encouraging evidence that the approach can support Universal Design (cognitive offloading without visual interfaces) and serve as a useful data-generation engine for grounded Vision and Language Navigation (VLN).


\section{Discussion \& Limitations}

Deployment in a real quasi-static environment reveals boundary conditions that motivate future work.

\textbf{Perception and FOV Limitations:} Mobile scanning remains vulnerable to field-of-view clipping. Items placed on extreme bottom or top shelves were occluded from the chest-level trajectory, causing localized search failures. While \emph{zone-level estimation} can mitigate some missing detections, robust long-term deployment will likely require improved scan protocols, active re-scanning, or incremental map updates.

\textbf{Semantic Aliasing in Localization:} The one-shot localizer achieves strong zone-level performance, but as demonstrated in Figure~\ref{fig:mirrored}, identical semantic arrangements viewed from mirrored positions induce perceptual aliasing. This makes the method better suited for global initialization as opposed to full tracking. Fusing semantic hypotheses with temporal odometry or depth cues is the most direct next step towards a full tracking solution. 

\textbf{Evaluation Scope:} The provided evaluation should be interpreted at the appropriate strength: the LLM-based study provides controlled, comparative evidence about instruction quality and the ecological probe provides encouraging real-world evidence, but neither substitute for larger-scale human-subject studies with target users and direct psychophysiological measures of cognitive load, confidence, and accessibility outcomes. 

\section{Safe and Responsible Innovation Statement}
GIST addresses critical ethical and societal considerations. By explicitly decoupling geometric safety from semantic VLM reasoning, our architecture lowers the physical risks of AI spatial hallucination, ensuring safer navigational guidance for vulnerable populations, including Blind and Low-Vision (B/LV) users. To preserve data privacy, GIST is designed to process consumer-grade mobile LiDAR and visual data locally, avoiding continuous cloud streaming of sensitive indoor environments. While the system enhances autonomy through universal design, ongoing care must be taken to mitigate representation bias in underlying foundation models.

\section{Conclusion}

In this work, we introduced \emph{GIST}, a framework that bridges the gap between raw metric geometry and high-level multimodal reasoning. By extracting an intelligent semantic topology from consumer-grade mobile scans, GIST provides a shared, lightweight representation that grounds spatial interaction in dense, quasi-static environments. Rather than relying on computationally heavy 3D maps or brittle, ungrounded visual sequences, our approach demonstrates that separating deterministic navigation structures from semantic AI reasoning yields highly robust spatial awareness.

We validated the versatility of this representation across a spectrum of downstream tasks essential for both human and embodied AI search and navigation. By leveraging this shared topology, GIST successfully enables intent-aware semantic search with graceful zone-level fallbacks, infrastructure-free global localization, and the synthesis of safe, landmark-rich routing instructions. Because GIST requires no external environmental infrastructure such as Bluetooth beacons or RFID tags, it is applicable to cluttered, quasi-static, uninstrumented environments including retail stores, warehouses, libraries, and hospitals. Ultimately, GIST establishes a scalable, resilient foundation for the next generation of multimodal spatial interaction systems.

\bibliographystyle{ACM-Reference-Format}
\bibliography{main}

\end{document}